\documentclass{article}

     \usepackage[nonatbib,preprint]{neurips_2021}

\usepackage[utf8]{inputenc} 
\usepackage[T1]{fontenc}    
\usepackage{hyperref}       
\usepackage{url}            
\usepackage{booktabs}       
\usepackage{amsfonts}       
\usepackage{nicefrac}       
\usepackage{microtype}      
\usepackage{amsmath}
\usepackage{amsthm}
\usepackage{amsthm}
\usepackage{bbm}
\usepackage{subcaption}
\usepackage{booktabs}
\usepackage{todonotes}

\usepackage{graphicx}

\newcommand{\eps}{\epsilon}
\newcommand{\set}[1]{\left\{#1\right\}}

\title{Evading Adversarial Example Detection Defenses with Orthogonal Projected Gradient Descent}

\author{%
  Oliver Bryniarski\thanks{Equal contributions. Authored alphabetically.} \\ UC Berkeley \And
  Nabeel Hingun$^*$ \\ UC Berkeley \And
  Pedro Pachuca$^*$ \\ UC Berkeley \And
  Vincent Wang$^*$ \\ UC Berkeley \And
  Nicholas Carlini \\ Google
}

\begin{document}

\maketitle

\begin{abstract}
  Evading adversarial example detection defenses requires finding adversarial examples that must simultaneously
  (a) be misclassified by the model and (b) be detected as non-adversarial.
  We find that existing attacks that attempt to satisfy multiple simultaneous constraints
  often over-optimize against one constraint at
  the cost of satisfying another.
  We introduce \emph{Orthogonal Projected Gradient Descent}, an improved attack technique to generate adversarial examples that avoids this problem by
  orthogonalizing the gradients
  when running standard gradient-based attacks.
  We use our technique to evade four state-of-the-art detection defenses,
  reducing their accuracy to 0\% while maintaining a 0\% detection rate.
\end{abstract} 

\section{Introduction}

Generating \emph{adversarial examples} \cite{szegedy2013intriguing,biggio2013evasion}, inputs designed by an adversary
to cause a neural network to behave incorrectly, is straightforward.
By performing
input-space gradient descent \cite{carlini2017towards,madry2017towards}, it is possible to maximize
the loss of arbitrary examples at test time.
This process is both efficient and highly effective. Despite great efforts by the community,
attempts at designing defenses against adversarial examples have been largely unsuccessful and gradient-descent attacks continue to circumvent new defenses, even those that attempt to make finding gradients difficult or impossible
\cite{athalye2018obfuscated,tramer2020}.

As a result, many defenses aim to make generating adversarial examples more difficult by requiring
additional constraints on inputs for them to be considered successful.
Defenses that rely on \emph{detection}, for example, will reject inputs if a secondary
detector model determines the input is adversarial \cite{metzen2017detecting,xu2017feature}.
Turning a benign input $x$ into an adversarial example 
$x'$ thus now requires  fooling both
the original classifier, $f$, \emph{and} the detector, $g$, simultaneously.

Traditionally, this is done by constructing a single loss function $\mathcal{L}$ that
jointly penalizes the loss on $f$ and the loss on $g$, 
e.g., by defining $\mathcal{L}(x') = \mathcal{L}(f) + \lambda \mathcal{L}(g)$ and then minimizing $\mathcal{L}(x')$ with gradient descent \cite{carlini2017adversarial}.
Unfortunately, many defenses which develop evaluations using this strategy have had limited success in evaluating this way---not only must $\lambda$ be tuned appropriately,
but the gradients of $f$ and $g$ must also be well behaved.
%

\textbf{Our contributions.}
We develop a new attack technique designed to construct adversarial examples
that simultaneously satisfy multiple constraints.
%
%
%
%
Our attack approach is a modification of standard gradient descent \cite{madry2017towards}
and requires changing just a few lines of code.
Given two objective functions $f$ and $g$,
instead of taking gradient descent steps that optimize the joint loss function $f + \lambda g$, we selectively take gradient descent steps
on either $f$ or $g$.
This makes our attack both simpler and easier to analyze than prior attack approaches.

We use our technique to evade four state-of-the-art and previously-unbroken defenses to adversarial examples: the Honeypot defense (CCS'20) \cite{shan2020gotta}, Dense Layer Analysis (IEEE Euro S\&P'20) \cite{Sperl2019}, Sensitivity Inconsistency Detector (AAAI'21) \cite{tian2021detecting}, and
the SPAM detector presented in Detection by Steganalysis (CVPR'19) \cite{liu2019detection}.
In all cases, we successfully reduce the accuracy of the protected classifier
to $0\%$ while maintaining a detection AUC of less than $0.5$---meaning the detector
performs worse than random guessing.

The code we used to produce the results in this paper is published on GitHub at the following URL: \url{https://github.com/v-wangg/OrthogonalPGD.git}.

\section{Background}

\subsection{Notation}
We consider classification neural networks $f : \mathbb{R}^d \to \mathbb{R}^n$ that receive a $d$-dimensional
input vector (in this paper, images) $x \in \mathbb{R}^d$ and output
an $n$-dimensional prediction vector $f(x) \in \mathbb{R}^n$.
We then use the notation $g : \mathbb{R}^d \to \mathbb{R}$ 
to denote some other constraint which must also be satisfied,
where $g(x) < 0$ when the constraint is satisfied and $g(x) > 0$ if it is violated. 
For detection defenses, this function $g$ is the detector and
higher values corresponding to higher likelihood of the input being an adversarial example.
\footnote{These definitions are without loss of generality. For example, a two-class detector $g$ can be converted to a one-class detector by subtracting $p_{adversarial}$ from $p_{benign}$.} We write $c(x)=y$ to say that the true label for input $x$ is the label $y$. When it is clear from the context, we abuse notation and write $y = f(x)$ to denote the arg-max most likely label under the model $f$. We use $\mathcal{L}$ to denote the loss for our classifier (e.g. cross entropy loss). 
Finally, we let $e(x)$ represent the embedding of an input $x$ at an intermediate
layer of $f$. Unless specified otherwise, $e$ returns the logit vector that immediately precedes the softmax activation. 

\subsection{Adversarial Examples}

Adversarial examples \cite{szegedy2013intriguing,biggio2013evasion} have been demonstrated in nearly every domain in which neural networks are used. \cite{adv_nlpAlzantot2018, adv_audioCW2018, advRLHuang2017} Given an input $x$ corresponding to label $c(x)$ and classifier $f$, an adversarial example is a perturbation $x'$ of the input such that $d(x, x') < \eps$ and $c(x') \ne t$ for some metric $d$.
The metric $d$ is most often that induced by a $p$-norm, typically either $||\cdot||_2$ or $||\cdot||_\infty$. With small enough perturbations under these metrics, the adversarial example $x'$ is not perceptibly different from the original input $x$.

\paragraph{Datasets.}
We attack each defense on the dataset that it performs best on.
All of our defenses operate on images.
For three of these defenses, this is the CIFAR-10 dataset \cite{Krizhevskycifar}, and for one, it is the ImageNet dataset \cite{deng2009imagenet}.
We constrain our adversarial examples for each paper under the threat model originally considered to perform a fair re-evaluation, but also generate adversarial examples with standard norms used extensively in prior work in order to make cross-defense evaluations meaningful.
We perform all evaluations on a single GPU. Our attacks on CIFAR-10 require just a few minutes, and for ImageNet a few hours.

\subsection{Detection Defenses}
\label{section:DetectionDefenses}

We focus our study on detection defenses. Rather than improve the robustness of the model to adversarial examples directly (e.g., through adversarial training \cite{madry2017towards} or certified approaches \cite{raghunathan2018certified,lecuyer2019certified,cohen2019certified}), detection defenses attempt to classify inputs as adversarial or benign \cite{metzen2017detecting,xu2017feature}. However, it is often possible to generate adversarial examples which simultaneously fool both the classifier and detector \cite{carlini2017adversarial}.
There have been several different strategies attempted to detect adversarial examples over the past
few years \cite{harnessing_adversarial, metzen2017detecting, feinman2017detecting, xu2017feature, meng2017magnet, ma2018characterizing, roth2019odds}.
Consistent with prior work, in this paper we work under the perfect-knowledge scenario: the adversary has direct access to both functions $f$ and $g$.




\subsection{Generating Adversarial Examples with Projected Gradient Descent}

Projected Gradient Descent \cite{madry2017towards} is a powerful first-order method for finding such adversarial examples.
Given a loss $\mathcal{L}(f, x, t)$ that takes a classifier, input, and desired target label, we optimize over the constraint set $S_\epsilon = \{z : d(x, z) < \epsilon\}$ and solve
\begin{equation}
     x' = \mathop{\text{arg min}}_{z \in S_\epsilon} \mathcal{L}(f, z, t) 
     \label{eqn:mainloss}
\end{equation}
by taking the following steps: $$x_{i+1} = \Pi_{S_\epsilon}\left(x_i -  \alpha \nabla_{x_i} \mathcal{L}(f, x_i, t) \right)$$ Here, $\Pi_{S_\epsilon}$ denotes projection onto the set $S_\epsilon$, and $\alpha$ is the step size. For example, the projection $\Pi_{S_\epsilon}(z)$ for $d(x,z) = ||x-z||_\infty$ is given by clipping $z$ to $[x-\epsilon, x+\epsilon]$. In this paper, we adapt PGD in order to solve optimization problems which involve minimizing multiple objective functions simultaneously. Wherever we describe gradient descent steps in later sections, we do not explicitly write $\Pi_{S_\epsilon}$ -- it is assumed that all steps are projected onto the constraint set.


\subsection{Related Attacks} 

Recent work has shown that it is possible to attack models with \emph{adaptive attacks} that target specific aspects of defenses. For detection defenses this process is often \emph{ad hoc}, involving alterations specific to each given defense \cite{tramer2020}. An independent line of work develops automated attack techniques that are reliable indicators of robustness \cite{hein2020}; however, in general, these attack approaches are difficult to apply to detection defenses. One useful output of our paper is a scheme that may help these automated tools evaluate detection defenses.


\section{Rethinking Adversarial Example Detection}

Before we develop our improved attack technique to break
adversarial example detectors, it will be useful
to understand why evaluating adversarial example detectors
is more difficult than evaluating standard classifiers.

Early work on adversarial examples often set up the problem slightly differently than we do above in Equation~\ref{eqn:mainloss}.
The initial formulation of an adversarial example \cite{szegedy2013intriguing,carlini2017towards} asks for
the smallest perturbation $\delta$ such that $f(x+\delta)$ is misclassified.
That is, these papers solved for
\[ \text{arg min}\, \lVert \delta \rVert_2\,\,\,\, \text{such that}\,\, f(x + \delta) \ne t \]
Solving this problem as stated is intractable.
It requires searching over a nonlinear constraint set, which is not feasible for
standard gradient descent.
As a result, these papers reformulate the search with the standard Lagrangian relaxation
\begin{equation}
    \text{arg min}\, \lVert \delta \rVert_2 + \lambda \mathcal{L}(f, x + \delta, t) 
    \label{eqn:old}
\end{equation}
This formulation is simpler, but still (a) requires tuning $\lambda$ to work well, and (b) is only guaranteed to be correct for convex functions $\mathcal{L}$---that it works for non-convex models like deep neural networks is not theoretically justified.
It additionally requires carefully constructing loss functions $\mathcal{L}$ \cite{carlini2017towards}.

Equation~\ref{eqn:mainloss} simplifies the setup considerably by just exchanging the constraint and objective.
Whereas in Equation~\ref{eqn:old} we search for the smallest perturbation that results in misclassification,
Equation~\ref{eqn:mainloss} instead finds an
input $x+\delta$ that maximizes the classifier's loss.
This is a simpler formulation because now the
constraint is convex, and so we can run standard gradient
descent optimization.

\textbf{Evading detection defenses is difficult because there are now two non-linear constraints}.
Not only must the input be constrained by a distortion bound and be misclassified by the base classifier, but we must \emph{also} have that they are not detected, i.e., with $g(x) < 0$.
This new requirement is nonlinear, and now it becomes impossible to side-step the problem by merely swapping the objective and the constraint as we did before: there will always be at
least one constraint that is a non-linear function, and so
standard gradient descent techniques can not directly apply.

In order to resolve this difficulty, the existing literature applies the same Lagrangian relaxation as was previously applied to constructing minimum-distortion adversarial examples.
That is, breaking a detection scheme involves solving
\begin{equation}
\mathop{\text{arg min}}_{x \in S_\epsilon} \mathcal{L}(f, x, t) + \lambda g(x) 
\label{eqn:breaklambda}
\end{equation}
where $\lambda$ is a hyperparameter that controls the relative
importance of fooling the classifier versus fooling the detector.
This formulation again brings back all of the reasons why the 
community moved past minimum-distortion adversarial examples.

\subsection{Perturbation Waste}
The fundamental failure mode for attacks on detection defenses that build on Equation~\ref{eqn:breaklambda} is what we call \emph{perturbation waste}.
Intuitively, we say that an adversarial example has wasted its perturbation budget if it has over-optimized against (for example) the detector so that $g(x)$ is well below 0 but so that it is still correctly classified.
More formally if 
an adversarial example $x'$ must satisfy two constraints
$c_1(x') \le 0 \wedge c_2(x') \le 0$ then we say it
has \emph{perturbation waste} if (without loss of generality)
$c_1(x') < -\alpha < 0$ but $c_2(x') > 0$.
We can now talk precisely about why generating adversarial examples that
break detection defenses through Equation~\ref{eqn:breaklambda} is
not always optimal: doing this often causes perturbation waste.

Consider a benign input pair $x$, a target label $t \ne c(x)$, and its corresponding (not yet known) adversarial example $x'$.
This input definitionally satisfies $f(x') = t$ and $g(x') < 0$.
Assuming the gradient descent search succeeds and optimizing Equation~\ref{eqn:breaklambda} reaches a global minimum,
we can derive upper and lower bounds on what the value of $\lambda$ should have been.
However, this range of acceptable $\lambda$ values is not going to be known ahead of time, and so requires additional search.
However, worse, there is a second set of constraints:
because the loss function is non-convex, the value
of $\lambda$ must be valid not only at the \emph{end} of optimization but also at the \emph{start} of optimization.
In the worst case this might introduce incompatibilities where no single value of $\lambda$ works throughout the generation process requiring tuning $\lambda$ during a single adversarial example search.

\section{Our Attack Approaches}

We now present our attack strategy designed to generate adversarial
examples that do not exhibit
perturbation waste.
We develop two related attack strategies that are designed to minimize
perturbation waste.
Then, in the following section we will apply these two attacks on defenses from the literature and show that they are indeed effective.

As we have been doing, each of our attack strategies defined below generates 
a targeted adversarial example $x'$ so that $f(x') = t$ but $g(x') < 0$.
Constructing an untargeted attack is nearly identical except for the substitution of maximization instead of minimization.

\subsection{Selective gradient descent}

Instead of minimizing the weighted sum of $f$ and $g$, 
our first attack completely eliminates the possibility for
perturbation waste by never optimizing
against a constraint once it becomes satisfied.
That is, we write our attack as
\begin{equation}
\mathcal{A}(x, t) = \mathop{\text{arg min}}_{x' : \lVert x - x' \rVert < \epsilon} 
\underbrace{\mathcal{L}(f, x', t) \cdot \mathbbm{1}[f(x) \ne t]  + g(x') \cdot \mathbbm{1}[f(x) = t]}_{\mathcal{L}_{\text{update}}(x, t)}.
\label{eqn:first}
\end{equation}
The idea here is that instead of minimizing a convex combination of the two loss functions,
we selectively optimize either $f$ or $g$ depending on if $f(x) = t$,
ensuring that updates are always helping to improve either the loss on $f$
or the loss on $g$.

Another benefit of this style is that it decomposes the gradient step into two updates,
which prevents \emph{imbalanced gradients}, where the gradients for two loss
functions are not of the same magnitude and result in unstable optimization \cite{jiang2020imbalanced}.
In fact, our loss function can be viewed directly in this lens as following the
margin decomposition proposal \cite{jiang2020imbalanced} by observing that
\begin{equation}
\nabla \mathcal{L}_{\text{update}}(x, t) = \begin{cases}
\nabla \mathcal{L}(f, x, t) & \text{if } f(x) \ne t \\
\nabla g(x) & \text{if } f(x) = t.
\end{cases}
\label{eqn:selective}
\end{equation}

That is, with each iteration, we either take gradients on $f$ or on $g$ depending
on whether $f(x) = t$ or not.
The equivalence can be shown by computing $\nabla \mathcal{L}(x)$ from Equation~\ref{eqn:first}.

\subsection{Orthogonal gradient descent}

The prior attack, while mathematically correct, might encounter numerical stability difficulties.
Often, the gradients of $f$ and $g$ point in opposite directions, that is,
$\nabla f \approx - \nabla g$.
As a result, every step spent optimizing $f$ causes backwards progress on
optimizing against $g$.
This results in the optimizer constantly ``undoing'' its own progress after
each step that is taken.
We address this problem by giving a slightly different update rule that again
will solve Equation~\ref{eqn:selective},
however this time by optimizing
\begin{equation}
\mathcal{L}_{\text{update}}(x, t) = \begin{cases}
\nabla \mathcal{L}(f, x, t) - \textrm{proj}_{\nabla \mathcal{L}(f, x, t)} \nabla g(x) & \text{if } f(x) \ne t \\
\nabla g(x) - \textrm{proj}_{\nabla g(x)} \nabla \mathcal{L}(f, x, t)& \text{if } f(x) = t.
\end{cases}
\label{eqn:orthogonal}
\end{equation}

Note that $ \nabla g(x) ^\perp = \nabla \mathcal{L}(f, x, t) - \textrm{proj}_{\nabla \mathcal{L}(f, x, t)} \nabla g(x)$ is orthogonal to the gradient $\nabla g(x)$, and similarly $\nabla \mathcal{L}(f, x, t)^\perp$ is orthogonal to $\nabla \mathcal{L}(f, x, t)$.
The purpose of this update is to take gradient descent steps with respect to one of $f$ or $g$ in such a way that we do not significantly disturb the loss of the function not chosen. 
In this way, we prevent our attack from taking steps that undo work done in
previous iterations of the attack.

\section{Case Studies}
\label{section:CaseStudies}

We validate the efficacy of our attack by using it to circumvent four previously unbroken, state-of-the-art 
defenses accepted at top computer security or machine learning venues. Three of the case study utilizes models and code obtained directly from their respective authors. In the final case the original authors provided us with matlab source code that was not easily used, which we re-implemented.


One factor we have not yet mentioned is that implicit to the setup of $g$ is a \emph{threshold} that adjusts the trade-off between true positives
and false positives.
Until now we have said that $g(x) < 0$ implies the input is classified as benign. However,
when considering alternate thresholds, we use the notation $\phi$ so that if $g(x) > \phi$, then $x$ is flagged as adversarial.
The choice of $\phi$ is made empirically as it determines the false positive rate of the detector--it is
up to the defender to choose an acceptable threshold depending on the situation.

\textbf{Attack Success Rate Definition.}
We evaluate the success of our attack by a metric that we call
attack success rate at N (SR@N for short).
SR@N is defined as the fraction of targeted attacks that succeed when the defense's false positive rate is set to $N\%$.
For example, a $94\%$ $SR@5$ could either be achieved through
$94\%$ of inputs being misclassified as the target class and $0\%$ being detected as adversarial, or by $100\%$ of inputs being misclassified as the target class and $6\%$ being detected as adversarial, or some combination thereof.
We report SR@5 and SR@50 for our main results.
The value $5\%$ is used in many prior defenses in the literature \cite{ma2018characterizing,xu2017feature}, and $50\%$ is an extreme upper bound and would reduce the model's accuracy by half.
We also give the full ROC curve of the detection rate for a more complete analysis.

Finally, note that all of our attacks are \emph{targeted} attacks where we choose the target uniformly at random from among the incorrect class labels. Untargeted attacks are in general an order of magnitude easier (because there are more possible incorrect labels).
We apply targeted attacks for the reasons listed in prior attack work \cite{athalye2018obfuscated}, primarily because if targeted attacks succeed,
then untargeted attacks certainly will.

\subsection{Honeypot Defense}

\begin{figure}

\begin{subfigure}[b]{.58\textwidth}
    \centering

    \begin{tabular}{l|rr|rr}
    \toprule
    
Attack & \multicolumn{2}{c}{eps=0.01} & \multicolumn{2}{c}{eps=0.031} \\
& SR@5 & SR@50 & SR@5 & SR@50 \\
\midrule
\cite{shan2020gotta} & 0.02 & - & - & - \\
\midrule
Orthogonal   & \textbf{1.0} & \textbf{0.93} & \textbf{1.0} & \textbf{0.92} \\
Selective   & 0.998 & 0.505 &  0.996 & 0.628 \\

\bottomrule
    \end{tabular}
    \vspace{2em}

    \caption{Attack success rate for our two proposed attacks.}
\end{subfigure}
\begin{subfigure}[b]{.48\textwidth}
    \includegraphics[scale=.6]{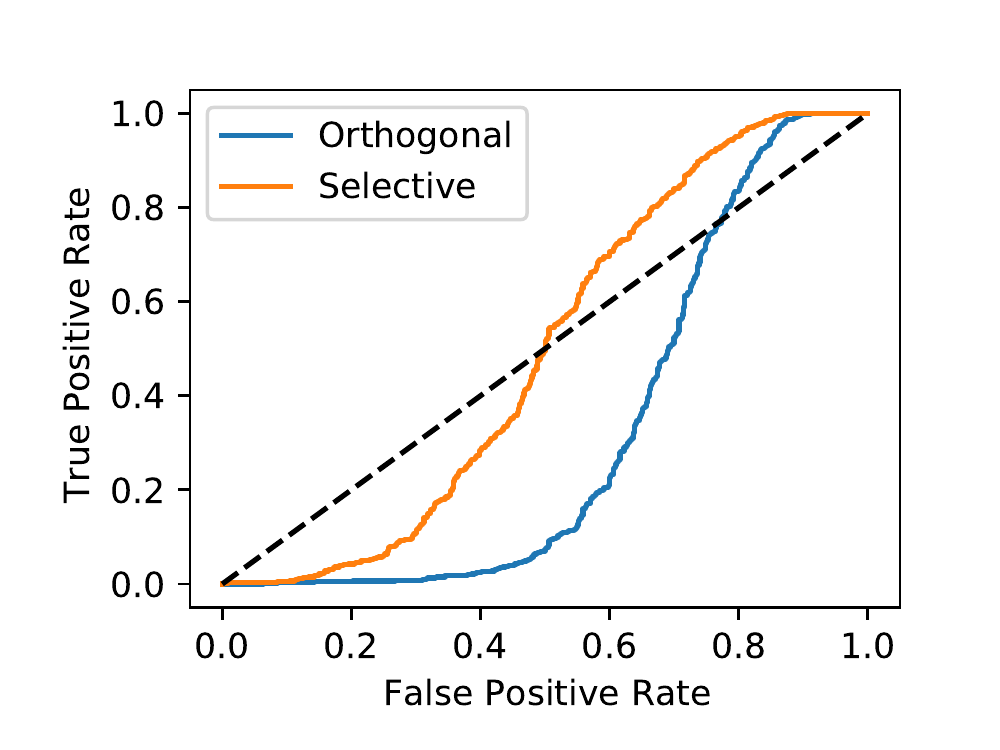}
    \caption{Defense ROC, $\varepsilon=0.01$}
\end{subfigure}
\caption{Honeypot attack evaluation. Compared to the originally reported 2\% success rate, our attack reaches a 100\% attack success rate under the same distortion bound. While the ROC curve does cross over the $x=y$ line, this only occurs after a FPR of $70\%$ which is completely unusable in practice.}
\label{fig:HoneypotFigure}
\end{figure}

The first paper we consider is the Honeypot Defense \cite{shan2020gotta}. Instead of preventing attackers from directly constructing adversarial examples, the authors propose to \emph{lure} attackers into producing specific perturbations that are easy to find and hard to ignore.
These perturbations are called ``honeypots'' or trapdoors and can be easily identified by a detector. For their evaluation on the MNIST and CIFAR-10 dataset, the authors use $5$ sets of randomly selected $3 \times 3$ squares per label.

Formally, consider an input $x$ to the classifier, $f$. During training, $f$ is injected with a honeypot, $\Delta$. The signature of a particular honeypot, $S_{\Delta}$, is the expectation of the neuron activations of $f$ over multiple sample inputs containing $\Delta$. During inference, the neuron activation pattern $e(x)$ is compared to $S_{\Delta}$ using cosine similarity. Specifically, for a predefined threshold $\phi$, if $\cos(e(x), S_{\Delta}) > \phi$, then $x$ is flagged as adversarial. One additional modification the authors make is to use neuron randomization. When creating the signatures, a random sample of neuron activations is ignored and the same set of neurons is ignored when calculating the cosine similarity during adversarial attack detection.

We test our method of generating adversarial images on CIFAR-10 with an $l_{\infty}$ bound of $\epsilon = \frac{8}{255}$. The original model has a baseline accuracy of $.87$. To test the defense, we generate targeted attacks choosing the target from among the incorrect labels. 
As a simple baseline attack, we confirm that FGSM achieves an accuracy of $.06$ on the classifier but the ROC AUC of the detector is $.994$. 

To evaluate the efficacy of our attack, we use the authors pre-trained code\footnote{https://github.com/Shawn-Shan/trapdoor}
and convert their models from TensorFlow to PyTorch to run our attacks.
We find that our attack is extremely effective and has an attack success rate
above $99.9\%$, even at a $50\%$ false positive rate.
In Table \ref{fig:HoneypotFigure} we give the attack results for several configurations of our attack.
We shared our results with the original defense authors who confirmed our adversarial examples successfully evaded their detection (an important step,
given that we had converted the model to PyTorch).

\subsection{Dense Layer Analysis}

Many recent defenses analyze the hidden activations of neural networks while processing benign and adversarial inputs (e.g., following \cite{metzen2017detecting}). These defenses aim to capitalize on differences in activation patterns among benign and adversarial inputs to train a separate classifier capable of detecting inputs as adversarial.

The most recent paper in this direction Sperl \emph{et al.} extract dense layer activation patterns among benign and adversarial inputs and train a secondary binary classifier that detects adversarial examples \cite{Sperl2019}. The authors do this by first performing a forward pass through a target neural network with both adversarial and benign inputs to create a mixed-feature dataset of activation-label pairs. Then, using the mixed-feature dataset, they train a secondary binary classifier capable of discerning between adversarial and benign inputs. When evaluating their models, the authors pass an input through the target model to obtain the activation feature vectors for a particular input as well as a potential classification. They then pass this feature vector through the secondary classifier. If the secondary classifier alerts that the input was adversarial, the classification is thrown away. Otherwise, classification proceeds as normal. 

Sperl \emph{et al.} evaluate this defense with 5 leading adversarial attacks on the MNIST and CIFAR-10 datasets using several models and report high accuracies for benign inputs and high detection rates for adversarial inputs. The authors report a worst-case individual attack accuracy of 0.739.


In accordance with our framework, we assign the cross entropy loss of the classifier to our primary function and binary cross entropy loss of the detector as our secondary function. 

\begin{figure}

\begin{subfigure}[b]{.58\textwidth}
    \centering

    \begin{tabular}{l|rr|rr}
    \toprule
    
Attack & \multicolumn{2}{c}{eps=0.01} & \multicolumn{2}{c}{eps=0.031} \\
& SR@5 & SR@50 & SR@5 & SR@50 \\
\midrule
\cite{sperl2020dla} & $\le$0.13$^*$ & - & - & -  \\
\midrule
Orthogonal   & 0.374 & 0.163 & \textbf{1.0} & 0.718 \\
Selective   & \textbf{0.83} & \textbf{0.441} & \textbf{1.0} & \textbf{0.865} \\
\bottomrule
    \end{tabular}
    \vspace{1em}

    \caption{Attack success rate for our two proposed attacks. $^*$The original paper did not report at $5\%$ FPR,
    the closest we could use was $13\%$ TPR at a 20\% FPR.
However our attack succeeds $83\%$ of the time even with a $4\times$ lower false positive rate.}
\end{subfigure}
\begin{subfigure}[b]{.48\textwidth}
    \includegraphics[scale=.6]{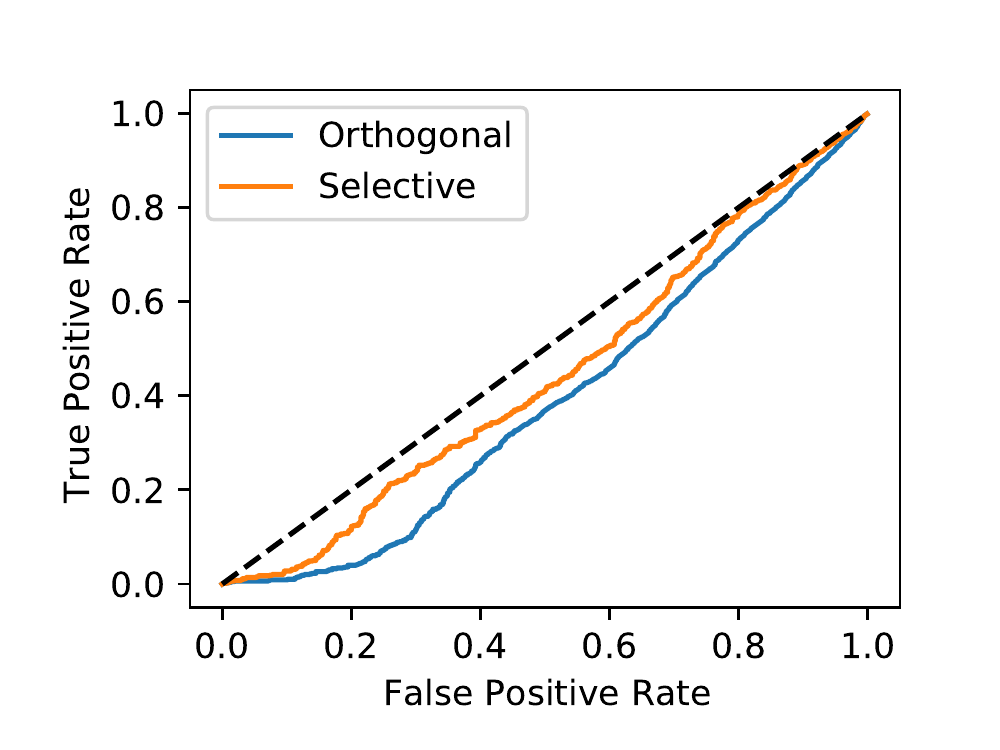}
    \caption{Defense ROC, $\varepsilon=0.01$}
\end{subfigure}
\caption{DLA attack evaluation. Our attack succeeds with $83\%$ probability compared to the original evaluation of $13\%$ (with $\varepsilon=0.01$), and $100\%$ of the time under the more typical $8/255$ constraint.
}
\label{fig:DLAFigure}
\end{figure}

We obtain source code and pre-trained defense models from the authors in order to ensure that our attack matches the defense as closely as possible.
We now detail the results of our attack at $\eps = .01$ and at $\eps = .03$ at false positive rates of 5\% and 50\% in Figure \ref{fig:DLAFigure}. We find that our attack is extremely effective, resulting in an accuracy of 0 at a detection rate of 0 with a false positive rate of 5\% under $\eps = .03$ bounds.
Finally, to validate that our attack succeeded, we again shared the resulting adversarial examples with the authors who confirmed our attack results.

\subsection{Sensitivity Inconsistency of Spatial-Transform Domain}

We next evaluated our attack on the Sensitivity Inconsistency Detector (SID) proposed by Tian \emph{et al.} \cite{tian2021detecting}. This defense relies on the observations of Fawzi \emph{et al.} \cite{Fawzi2018} that adversarial examples are movements, in the form of perturbations, of benign inputs in a decision space along an adversarial direction. Tian \emph{et al.} then conjecture that, because adversarial examples are likely to lie near highly-curved decision boundaries, and benign inputs lie away from such boundaries, fluctuations in said boundaries will often result in a change in classification of adversarial examples but not in classification of benign inputs.

To measure sensitivity against decision boundary transformations, Tian \emph{et al.} design a dual classifier which is the composition of a weighted additive wavelet transform layer and a DNN classifier with the same structure as the original classifier. When doing a forward pass of the system, the authors run an input through both the primal and the dual model, then pass both results to the detector that discriminates among adversarial and benign classes. With these models, the authors then define their so-called feature of sensitivity inconsistency $S(x_0)$.
\[S(x_0) = \{ f_i(x_0) - g_i(x_0) \}_{i=1}^K \]
where $f_i(x_0)$ and $g_i(x_0)$ are the predictions of the primal and the dual respectively. Input $x_0$ is  classified as adversarial if $S(x_0)$ is greater than a threshold $\phi$. SID achieves improved adversarial example detection performance, especially in cases with small perturbations in inputs. The authors report a worst-case, individual attack detection AUC \% of 0.95.

\begin{figure}

\begin{subfigure}[b]{.58\textwidth}
    \centering
    \begin{tabular}{l|rr|rr}
    \toprule
    
Attack & \multicolumn{2}{c}{eps=0.01} & \multicolumn{2}{c}{eps=0.031} \\
& SR@5 & SR@50 & SR@5 & SR@50 \\
\midrule
\cite{tian2021detecting} & $\le$ 0.09$^*$ & - & - & - \\
\midrule
Orthogonal   & \textbf{0.931} & \textbf{0.766} & \textbf{1.0} & \textbf{0.984} \\
Selective   & 0.911 & 0.491& \textbf{1.0} & 0.886 \\
\bottomrule
    \end{tabular}
    \vspace{1em}

    \caption{Attack success rate for our two proposed attacks. 
    $^*$The original paper only reports AUC values and does not report true positive/false positive rates. The value of $9\%$ was obtained by running PGD on the author's defense implementation.}
\end{subfigure}
\begin{subfigure}[b]{.48\textwidth}
    \includegraphics[scale=.6]{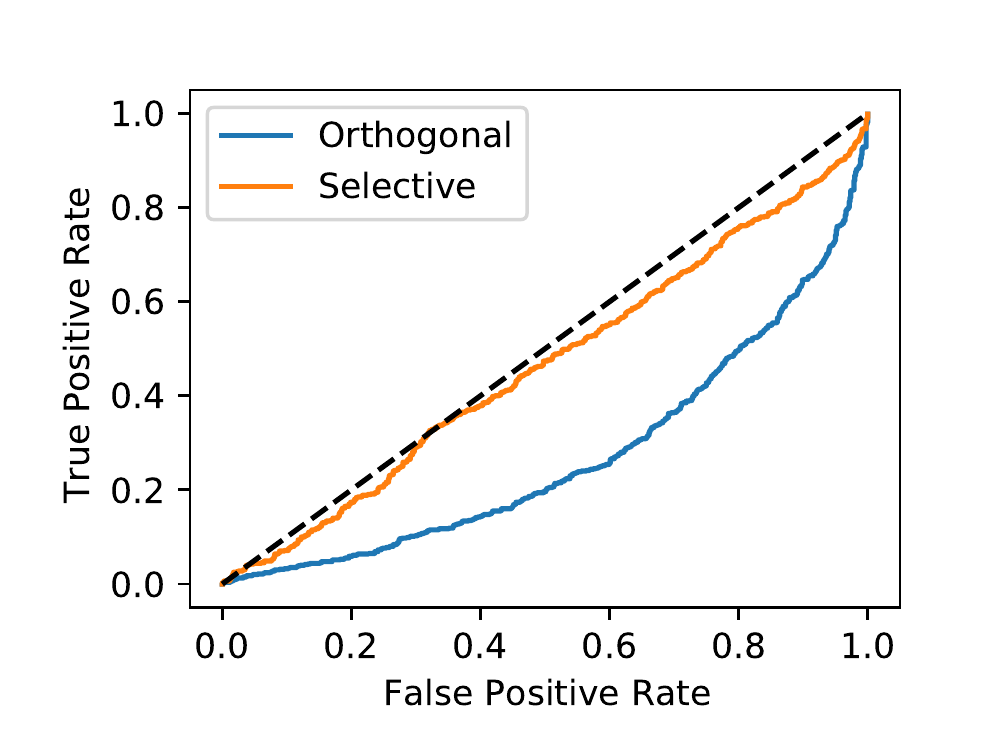}
    \caption{Defense ROC, $\varepsilon=0.01$}
\end{subfigure}
\caption{SID attack evaluation. Our attack succeeds with $91\%$ probability compared to the original evaluation of $9\%$ under a $\varepsilon=0.01$-norm constraint.}
\label{fig:SIDFigure}
\end{figure}

Now, we want to create adversarial examples that are misclassified by the original model and not flagged as adversarial by the Sensitivity Inconsistency Detector. We assign the loss of our target model to our primary function and the loss of the Sensitivity Inconsistency Detector as our secondary function. The initial target model had an accuracy of .94 and deemed .06 of all inputs adversarial.

We again obtain source code from the authors along with pre-trained models to ensure evaluation correctness.
We describe our attack's results at $\eps = .01$ and at $\eps = .03$ at false positive rates of 5\% and 50\% in Figure \ref{fig:SIDFigure}. Our attack works well in this case and induces an accuracy of 0 at a detection rate of 0 with a false positive rate of 5\% under $\eps = .03$ bounds.

\subsection{Detection through Steganalysis}

Since adversarial perturbations alter the dependence between pixels in an image, Liu \emph{et al.} \cite{liu2019detection} propose a defense which uses a \emph{steganalysis}-inspired approach to detect ``hidden features'' within an image. These features are then used to train binary classifiers to detect the perturbations. Unlike the prior defenses, this paper evaluates on ImageNet, reasoning that small images such as those from CIFAR-10 and MNIST do not provide enough inter-pixel dependency samples to construct efficient features for adversarial detection, so we attack this defense on ImageNet.

As a baseline, the authors use two feature extraction methods: SPAM and Spatial Rich Model. For each pixel $X_{i, j}$ of an image $X$, SPAM takes the difference between adjacent pixels along 8 directions. For the rightward direction, a difference matrix $A^{\rightarrow}$ is computed so that $A_{i, j}^{\rightarrow} = X_{i, j} - X_{i, j+1}$. A transition probability matrix $M^{\rightarrow}$ between pairs of differences can then be computed with
\[
    M^{\rightarrow}_{x, y} = Pr(A_{i, j+1}^{\rightarrow} = x | A_{i, j}^{\rightarrow} = y)
\]
where $x, y \in \set{-T, ..., T}$, with $T$ being a parameter used to control the dimensionality of the final feature set $F$. We use $T=3$ in accordance with that used by the authors. The features themselves are calculated by concatenating the average of the non-diagonal matrices with the average of the diagonal matrices:
\[
    F_{1, ..., k} = \frac{M^{\rightarrow} + M^{\leftarrow} + M^{\uparrow} + M^{\downarrow}}{4}
\qquad
        F_{k+1, ..., 2k} = \frac{M^{\rightarrow} + M^{\leftarrow} + M^{\uparrow} + M^{\downarrow}}{4}
\]
In order to use the same attack implementation across all defenses, we reimplemented this defense in PyTorch (the authors implementation was in matlab). Instead of re-implementing the full FLD ensemble \cite{kodovsky2011ensemble} used by the authors, we train a 3-layer fully connected neural network on SPAM features and use this as the detector. This allows us to directly investigate the claim that SPAM features can be reliably used to detect adversarial examples, as FLD is a highly non-diiferentiable operation and is not a fundamental component of the defense proposal.

The paper also proposes a second feature extraction method named ``Spatial Rich Model'' (SRM) that we do not evaluate against. This scheme follows the same fundamental principle as SPAM in modeling inter-pixel dependencies---there is only a marginal benefit from using these more complex models, and so we analyze the simplest variant of the scheme.

\begin{figure}

\begin{subfigure}[b]{.58\textwidth}
    \centering
    \begin{tabular}{l|rr|rr}
    \toprule
    
Attack & \multicolumn{2}{c}{eps=0.01} & \multicolumn{2}{c}{eps=0.031} \\
& SR@5 & SR@50 & SR@5 & SR@50 \\
\midrule
\cite{liu2019detection} & 0.03 & - & .03 & -  \\
\midrule
Orthogonal   & \textbf{0.988} & \textbf{0.54} & \textbf{1.0} & \textbf{0.62}  \\
\bottomrule
    \end{tabular}
    \vspace{1em}

    \caption{Attack success rate for our proposed attack.
    For computational efficiency, we only run our Orthogonal attack
    as the detection model has a throughput of one image per second.}
\end{subfigure}
\begin{subfigure}[b]{.48\textwidth}
    \includegraphics[scale=.6]{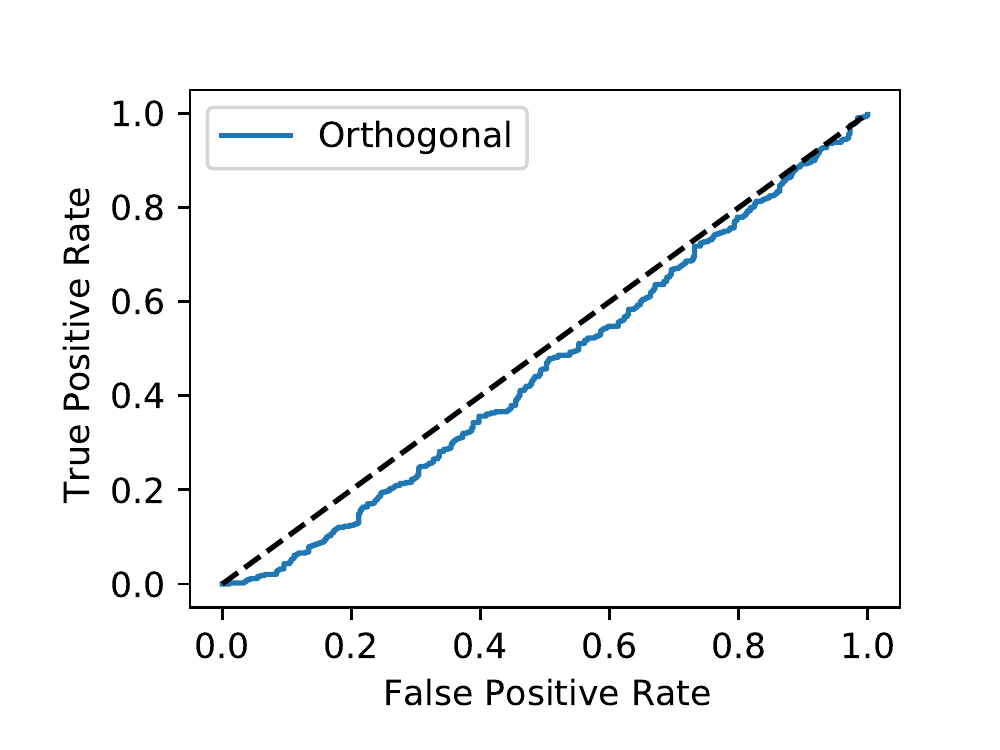}
    \caption{Defense ROC, $\varepsilon=0.01$}
\end{subfigure}
\caption{Steganalysis attack evaluation. We find it difficult to decrease the
detection score lower than the original score on the non-adversarial input, thus the
AUC is almost exactly 0.5.}
\label{fig:SteganalysisFigure}
\end{figure}

Notice that SPAM requires the difference matrices $A$ to be discretized in order for the dimensionality of the transition probability matrices $M$ to be finite. To make this discretization step differentiable and compatible with our attacks, we define a count matrix $X$ where, for example, $X_{x,y}^\rightarrow$ counts, for any every pair $i, j$, the number of occurrences of $y$ in $A_{i, j}^\rightarrow$ and $x$ in $A_{i, j+1}^\rightarrow$. $M^\rightarrow_{x,y}$ is then defined by:
\begin{equation*}
    M_{x,y}^\rightarrow = P(A_{i, j+1}^\rightarrow = x | A_{i, j}^\rightarrow = y) = \frac{X_{x, y}^\rightarrow}{\sum_{x'}X_{x',y}^\rightarrow}
\end{equation*}

To construct a differentiable approximation, consider without loss of generality the rightward difference matrix $A_1^{\rightarrow}$ for an image. We construct a shifted copy of it $A_2^{\rightarrow}$ so that $A_{2_{i, j}}^{\rightarrow} = A_{1_{i, j+1}}^{\rightarrow}$. We then define a mask $K$ so that 
\begin{align*}
    K_{i, j} = \mathbbm{1}[x \leq A_{2_{i, j}}^{\rightarrow} < x + 1 \cap y \leq A_{1_{i, j}}^{\rightarrow} < y + 1]
\end{align*}
Each element of the intermediate matrix $X_{x,y}^\rightarrow$ counts the number of pairs in $A_1^\rightarrow$ and $A_2^\rightarrow$ which would be rounded to $x$ and $y$ respectively after discretization:
\begin{align*}
    X_{x,y}^\rightarrow = \frac{\sum_{i, j}\left(K \circ A_2^\rightarrow\right)_{i, j}}{x}
\end{align*}
where $\circ$ is the Hadamard product. If we normalize $X^\rightarrow$ so that the sum of elements in each column is equal to $1$, we get the probability of difference values $x \in A_2^\rightarrow$ conditioned on column $y \in A_1^\rightarrow$. Thus, for any pair of indices $i, j$,
\begin{align*}
    M_{x,y}^\rightarrow = P(A_{2_{i, j}}^\rightarrow = x | A_{1_{i, j}}^\rightarrow = y) = \frac{X_{x, y}^\rightarrow}{\sum_{x'}X_{x',y}^\rightarrow}
\end{align*}
Using this differentiable formulation of SPAM feature extraction, we train an auxillary detector as described above and use its gradients to apply our attack on the original, non-differentiable detector.

The authors evaluate their defense on 4 adversarial attacks and report high accuracy for benign inputs and high detection rates for adversarial inputs. The best attack they develop still has a success rate less than $3\%$. In contrast, our attack on SPAM using the differentiable approximation has a success rate of $98.8\%$ when considering a $5\%$ false positive rate, with an AUC again less than the random guessing threshold of $0.5$.

\section{Conclusion}
\label{section:conclusion}

Generating adversarial examples that satisfy multiple constraints simultaneously (e.g., requiring that an input is both misclassified and deemed non-adversarial)
requires more care than generating adversarial examples that satisfy only one constraint (e.g., requiring only that an input is misclassified).
We find that prior attacks unnecessarily over-optimizes one constraint when another
constraint has not yet been satisfied.

Our new attack methodology of orthogonal and selective gradient descent prevent perturbation waste,
and ensure that the available perturbation budget is always ``spent'' on optimizing
the terms that are strictly necessary.
Our attack reduces the accuracy of four previously-unbroken detection methods to $0\%$ accuracy while maintaining a $0\%$ detection rate at $5\%$ false positive rates.

We believe our attack approach is generally useful.
For example, we believe that automated attack tools \cite{hein2020} would 
benefit from adding our optimization trick to their collection of known
techniques that could compose with other attacks.
However, we discourage future work from blindly applying this attack
without properly understanding its design criteria.
While this attack does eliminate perturbation waste for the defenses we
consider, it is not the only way to do so, and may not be the correct way
to do so in future defense evaluations.
Evaluating adversarial example defenses will necessarily require adapting
any attack strategies to the defense's design.

\section*{Acknowledgements}
We thank the authors of the papers we use in the case studies, who helped us answer questions specific to their respective defenses and agreed to share their code with us. We are also grateful to Alex Kurakin for comments on a draft of this paper.

\bibliography{paper}

\newcommand{\etalchar}[1]{$^{#1}$}
\begin{thebibliography}{FMDFS18}

\bibitem[ACW18]{athalye2018obfuscated}
Anish Athalye, Nicholas Carlini, and David Wagner.
\newblock Obfuscated gradients give a false sense of security: Circumventing
  defenses to adversarial examples.
\newblock In {\em International Conference on Machine Learning}, 2018.

\bibitem[ASE{\etalchar{+}}18]{adv_nlpAlzantot2018}
Moustafa Alzantot, Yash Sharma, Ahmed Elgohary, Bo{-}Jhang Ho, Mani~B.
  Srivastava, and Kai{-}Wei Chang.
\newblock Generating natural language adversarial examples.
\newblock {\em CoRR}, abs/1804.07998, 2018.

\bibitem[BCM{\etalchar{+}}13]{biggio2013evasion}
Battista Biggio, Igino Corona, Davide Maiorca, Blaine Nelson, Nedim
  {\v{S}}rndi{\'c}, Pave\~l Laskov, Giorgio Giacinto, and Fabio Roli.
\newblock Evasion attacks against machine learning at test time.
\newblock In {\em Joint European conference on machine learning and knowledge
  discovery in databases}, pages 387--402. Springer, 2013.

\bibitem[CH20]{hein2020}
Francesco Croce and Matthias Hein.
\newblock Reliable evaluation of adversarial robustness with an ensemble of
  diverse parameter-free attacks.
\newblock In {\em Proceedings of the 37th International Conference on Machine
  Learning}, pages 2206--2216. PMLR, 2020.

\bibitem[CRK19]{cohen2019certified}
Jeremy~M Cohen, Elan Rosenfeld, and J~Zico Kolter.
\newblock Certified adversarial robustness via randomized smoothing.
\newblock {\em arXiv preprint arXiv:1902.02918}, 2019.

\bibitem[CW17a]{carlini2017adversarial}
Nicholas Carlini and David Wagner.
\newblock Adversarial examples are not easily detected: Bypassing ten detection
  methods.
\newblock In {\em Proceedings of the 10th ACM Workshop on Artificial
  Intelligence and Security}, pages 3--14, 2017.

\bibitem[CW17b]{carlini2017towards}
Nicholas Carlini and David Wagner.
\newblock Towards evaluating the robustness of neural networks.
\newblock In {\em 2017 IEEE symposium on security and privacy}, pages 39--57.
  IEEE, 2017.

\bibitem[CW18]{adv_audioCW2018}
Nicholas Carlini and David Wagner.
\newblock Audio adversarial examples: Targeted attacks on speech-to-text.
\newblock In {\em 2018 IEEE Security and Privacy Workshops (SPW)}, pages 1--7,
  2018.

\bibitem[DDS{\etalchar{+}}09]{deng2009imagenet}
Jia Deng, Wei Dong, Richard Socher, Li-Jia Li, Kai Li, and Li~Fei-Fei.
\newblock Imagenet: A large-scale hierarchical image database.
\newblock In {\em 2009 IEEE conference on computer vision and pattern
  recognition}, pages 248--255. Ieee, 2009.

\bibitem[FCSG17]{feinman2017detecting}
Reuben Feinman, Ryan~R Curtin, Saurabh Shintre, and Andrew~B Gardner.
\newblock Detecting adversarial samples from artifacts.
\newblock {\em arXiv preprint arXiv:1703.00410}, 2017.

\bibitem[FMDFS18]{Fawzi2018}
Alhussein Fawzi, Seyed-Mohsen Moosavi-Dezfooli, Pascal Frossard, and Stefano
  Soatto.
\newblock Empirical study of the topology and geometry of deep networks.
\newblock In {\em Proceedings of the IEEE Conference on Computer Vision and
  Pattern Recognition (CVPR)}, June 2018.

\bibitem[GSS15]{harnessing_adversarial}
Ian Goodfellow, Jonathon Shlens, and Christian Szegedy.
\newblock Explaining and harnessing adversarial examples.
\newblock {\em International Conference on Learning Representations}, 2015.

\bibitem[HPG{\etalchar{+}}17]{advRLHuang2017}
Sandy~H. Huang, Nicolas Papernot, Ian~J. Goodfellow, Yan Duan, and Pieter
  Abbeel.
\newblock Adversarial attacks on neural network policies.
\newblock {\em CoRR}, abs/1702.02284, 2017.

\bibitem[JMW{\etalchar{+}}20]{jiang2020imbalanced}
Linxi Jiang, Xingjun Ma, Zejia Weng, James Bailey, and Yu-Gang Jiang.
\newblock Imbalanced gradients: A new cause of overestimated adversarial
  robustness.
\newblock {\em arXiv preprint arXiv:2006.13726}, 2020.

\bibitem[KFH12]{kodovsky2011ensemble}
Jan Kodovsky, Jessica Fridrich, and Vojtěch Holub.
\newblock Ensemble classifiers for steganalysis of digital media.
\newblock In {\em IEEE Transactions on Information Forensics and Security},
  pages 432--444, 2012.

\bibitem[KH09]{Krizhevskycifar}
A.~Krizhevsky and G.~Hinton.
\newblock Learning multiple layers of features from tiny images.
\newblock {\em Master's thesis, Department of Computer Science, University of
  Toronto}, 2009.

\bibitem[LAG{\etalchar{+}}19]{lecuyer2019certified}
Mathias Lecuyer, Vaggelis Atlidakis, Roxana Geambasu, Daniel Hsu, and Suman
  Jana.
\newblock Certified robustness to adversarial examples with differential
  privacy.
\newblock In {\em 2019 IEEE Symposium on Security and Privacy (SP)}, pages
  656--672. IEEE, 2019.

\bibitem[LZZ{\etalchar{+}}19]{liu2019detection}
Jiayang Liu, Weiming Zhang, Yiwei Zhang, Dongdong Hou, Yujia Liu, Hongyue Zha,
  and Nenghai Yu.
\newblock Detection based defense against adversarial examples from the
  steganalysis point of view.
\newblock In {\em Proceedings of the IEEE/CVF Conference on Computer Vision and
  Pattern Recognition}, pages 4825--4834, 2019.

\bibitem[MC17]{meng2017magnet}
Dongyu Meng and Hao Chen.
\newblock Magnet: a two-pronged defense against adversarial examples.
\newblock In {\em Proceedings of the 2017 ACM SIGSAC conference on computer and
  communications security}, pages 135--147, 2017.

\bibitem[MGFB17]{metzen2017detecting}
Jan~Hendrik Metzen, Tim Genewein, Volker Fischer, and Bastian Bischoff.
\newblock On detecting adversarial perturbations.
\newblock {\em arXiv preprint arXiv:1702.04267}, 2017.

\bibitem[MLW{\etalchar{+}}18]{ma2018characterizing}
Xingjun Ma, Bo~Li, Yisen Wang, Sarah~M Erfani, Sudanthi Wijewickrema, Grant
  Schoenebeck, Dawn Song, Michael~E Houle, and James Bailey.
\newblock Characterizing adversarial subspaces using local intrinsic
  dimensionality.
\newblock {\em arXiv preprint arXiv:1801.02613}, 2018.

\bibitem[MMS{\etalchar{+}}17]{madry2017towards}
Aleksander Madry, Aleksandar Makelov, Ludwig Schmidt, Dimitris Tsipras, and
  Adrian Vladu.
\newblock Towards deep learning models resistant to adversarial attacks.
\newblock {\em International Conference on Learning Representations}, 2017.

\bibitem[RKH19]{roth2019odds}
Kevin Roth, Yannic Kilcher, and Thomas Hofmann.
\newblock The odds are odd: A statistical test for detecting adversarial
  examples.
\newblock In {\em International Conference on Machine Learning}, pages
  5498--5507. PMLR, 2019.

\bibitem[RSL18]{raghunathan2018certified}
Aditi Raghunathan, Jacob Steinhardt, and Percy Liang.
\newblock Certified defenses against adversarial examples.
\newblock {\em arXiv preprint arXiv:1801.09344}, 2018.

\bibitem[SKC{\etalchar{+}}20]{sperl2020dla}
Philip Sperl, Ching-Yu Kao, Peng Chen, Xiao Lei, and Konstantin B{\"o}ttinger.
\newblock Dla: Dense-layer-analysis for adversarial example detection.
\newblock In {\em 2020 IEEE European Symposium on Security and Privacy
  (EuroS\&P)}, pages 198--215. IEEE, 2020.

\bibitem[SKCB19]{Sperl2019}
Philip Sperl, Ching{-}yu Kao, Peng Chen, and Konstantin B{\"{o}}ttinger.
\newblock {DLA:} dense-layer-analysis for adversarial example detection.
\newblock {\em CoRR}, abs/1911.01921, 2019.

\bibitem[SWW{\etalchar{+}}20]{shan2020gotta}
Shawn Shan, Emily Wenger, Bolun Wang, Bo~Li, Haitao Zheng, and Ben~Y Zhao.
\newblock Gotta catch'em all: Using honeypots to catch adversarial attacks on
  neural networks.
\newblock In {\em Proceedings of the 2020 ACM SIGSAC Conference on Computer and
  Communications Security}, pages 67--83, 2020.

\bibitem[SZS{\etalchar{+}}14]{szegedy2013intriguing}
Christian Szegedy, Wojciech Zaremba, Ilya Sutskever, Joan Bruna, Dumitru Erhan,
  and Rob Goodfellow, Ian an\ d~Fergus.
\newblock Intriguing properties of neural networks.
\newblock In {\em International Conference on Learning Representations (ICLR)},
  2014.

\bibitem[TCBM20]{tramer2020}
Florian Tram{\`{e}}r, Nicholas Carlini, Wieland Brendel, and Aleksander Madry.
\newblock On adaptive attacks to adversarial example defenses.
\newblock {\em CoRR}, abs/2002.08347, 2020.

\bibitem[TZLD21]{tian2021detecting}
Jinyu Tian, Jiantao Zhou, Yuanman Li, and Jia Duan.
\newblock Detecting adversarial examples from sensitivity inconsistency of
  spatial-transform domain.
\newblock {\em arXiv preprint arXiv:2103.04302}, 2021.

\bibitem[XEQ17]{xu2017feature}
Weilin Xu, David Evans, and Yanjun Qi.
\newblock Feature squeezing: Detecting adversarial examples in deep neural
  networks.
\newblock {\em arXiv preprint arXiv:1704.01155}, 2017.

\end{thebibliography}
\bibliographystyle{alpha}

\end{document}